\documentclass[10pt,twocolumn,letterpaper]{article}

 \usepackage{cvpr}              

\usepackage{graphicx}
\usepackage{subcaption}
\usepackage[table,xcdraw]{xcolor}
\usepackage{multirow}
\usepackage{makecell}
\definecolor{cvprblue}{rgb}{0.21,0.49,0.74}
\usepackage[pagebackref,breaklinks,colorlinks,allcolors=cvprblue]{hyperref}

\title{Knowledge-Guided Textual Reasoning for \\Explainable Video Anomaly Detection via LLMs}

\author{Hari Lee \\ 
}

\begin{document}
\maketitle
\begin{abstract}

We introduce Text-based Explainable Video Anomaly Detection (TbVAD), a language-driven framework for Weakly Supervised Video Anomaly Detection (WSVAD) that performs anomaly detection and explanation entirely within the textual domain.
Unlike conventional WSVAD models that rely on explicit visual features, TbVAD represents video semantics through language, enabling interpretable and knowledge-grounded reasoning.
The framework operates in three stages:
(1) transforming video content into fine-grained captions using a Vision-Language Model (VLM),
(2) constructing structured knowledge by organizing the captions into four semantic slots—action, object, context, and environment, and
(3) generating slot-wise explanations that reveal which semantic factors contribute most to the anomaly decision.
We evaluate TbVAD on two public benchmarks, UCF-Crime and XD-Violence, demonstrating that textual knowledge reasoning provides interpretable and reliable anomaly detection for real-world surveillance scenarios.


\end{abstract}

\section{Introduction}
\label{sec:introduction}

Video Anomaly Detection (VAD) is a critical task in computer vision, particularly for surveillance and public safety applications~\cite{Chengwei,sultani}.
While recent deep learning approaches have achieved remarkable progress, most existing methods still depend heavily on visual features extracted from raw frames or pretrained networks.
However, in practical surveillance scenarios with low-resolution CCTV footage, visual inputs often fail to capture subtle but meaningful cues such as minor object movements, unusual interactions, or small environmental changes.

Beyond raw visual signals, language offers a valuable yet underexplored representation of visual scenes~\cite{TEVAD,Zanella,girdhar,Du}.
Linguistic descriptions provide compact, interpretable, and generalizable semantics that align more closely with human reasoning.
We argue that, when combined with structured reasoning, textual representations can serve not only as an alternative modality for anomaly detection but also as a foundation for explainable reasoning that clarifies why an event is considered abnormal.

To this end, we propose Text-based Explainable Video Anomaly Detection (TbVAD) — a framework that performs anomaly detection and explanation entirely within the textual domain.
TbVAD integrates three complementary components:
(1) fine-grained captions generated by a Vision-Language Model (VLM) to describe local visual events;
(2) structured knowledge constructed via Large Language Model (LLM)–based multi-aspect summarization, capturing normal and abnormal behavioral patterns across four semantic slots — action, object, context, and environment; and
(3) slot-wise explanation generation, which interprets anomaly decisions based on the most influential semantic factors.
Together, these components form a comprehensive reasoning pipeline where domain-level priors and instance-specific descriptions jointly contribute to interpretable decision-making.

During inference, TbVAD computes slot-level importance scores to determine which semantic dimensions most influence the anomaly prediction.
Each caption is aligned with its corresponding structured knowledge, and the most relevant textual evidences are retrieved accordingly.
By integrating these evidences with fine-grained captions, TbVAD employs a lightweight language model to produce concise, human-understandable explanations for detected anomalies.
This unified design bridges the gap between quantitative detection and qualitative interpretation, enabling interpretable and knowledge-grounded VAD.

Finally, extensive experiments on UCF-Crime and XD-Violence datasets demonstrate that combining structured textual knowledge with fine-grained descriptions significantly enhances both accuracy and interpretability, validating the effectiveness of TbVAD’s text-based reasoning approach.

\section{Related Work}
\label{sec:relatedwork}
\subsection{Weakly Supervised Video Anomaly Detection}

Weakly supervised video anomaly detection (WSVAD) detects anomalies using video-level labels, eliminating the need for frame-level or instance-level annotations. That is, labels indicate whether a video contains an anomaly but do not specify the precise time or region where it occurs. This significantly reduces annotation costs, making large-scale surveillance datasets more feasible to construct. However, the coarse nature of supervision introduces label ambiguity, as normal and abnormal segments are mixed within a single video.

To handle this, multiple instance learning (MIL) has emerged as the dominant paradigm for WSVAD~\cite{sultani, zhong, MIL}. In this setting, a video is treated as a bag of temporal snippets, and the objective is to identify anomalous instances within positively labeled bags. Most approaches follow a two-stage pipeline: they first extract visual features using pretrained models and then train anomaly detectors on the snippet-level embeddings. This setup is efficient but often lacks semantic depth, especially in complex real-world scenarios.

To mitigate ambiguity and improve snippet-level discrimination, several refinements have been proposed. Zhong et al.~\cite{zhong} framed the problem as learning with noisy labels, employing alternate optimization strategies. Lv et al.~\cite{MIL} introduced a margin-based loss and higher-order temporal reasoning modules, while other methods like top-k instance mining~\cite{tian} and multiple sequence learning~\cite{li} aim to better capture localized anomalies by leveraging temporal continuity and snippet ordering.

Further advances incorporate strategies such as contrastive learning~\cite{Zhang} to separate normal and abnormal embeddings in feature space, self-training~\cite{yu_bmvc} to iteratively refine pseudo-labels, and scene-aware representations~\cite{cao_eccv, Singh_aaai} to reduce false positives by modeling background context.

Despite notable progress, WSVAD remains fundamentally constrained by its reliance on visual features and ambiguous supervision. In challenging environments—such as low-resolution or cluttered surveillance footage—these limitations can hinder the accurate detection of subtle anomalies. This motivates the exploration of alternative modalities that provide higher-level semantic abstraction and interpretability.

\begin{figure}[t]
  \centering
  \includegraphics[width=1\linewidth]{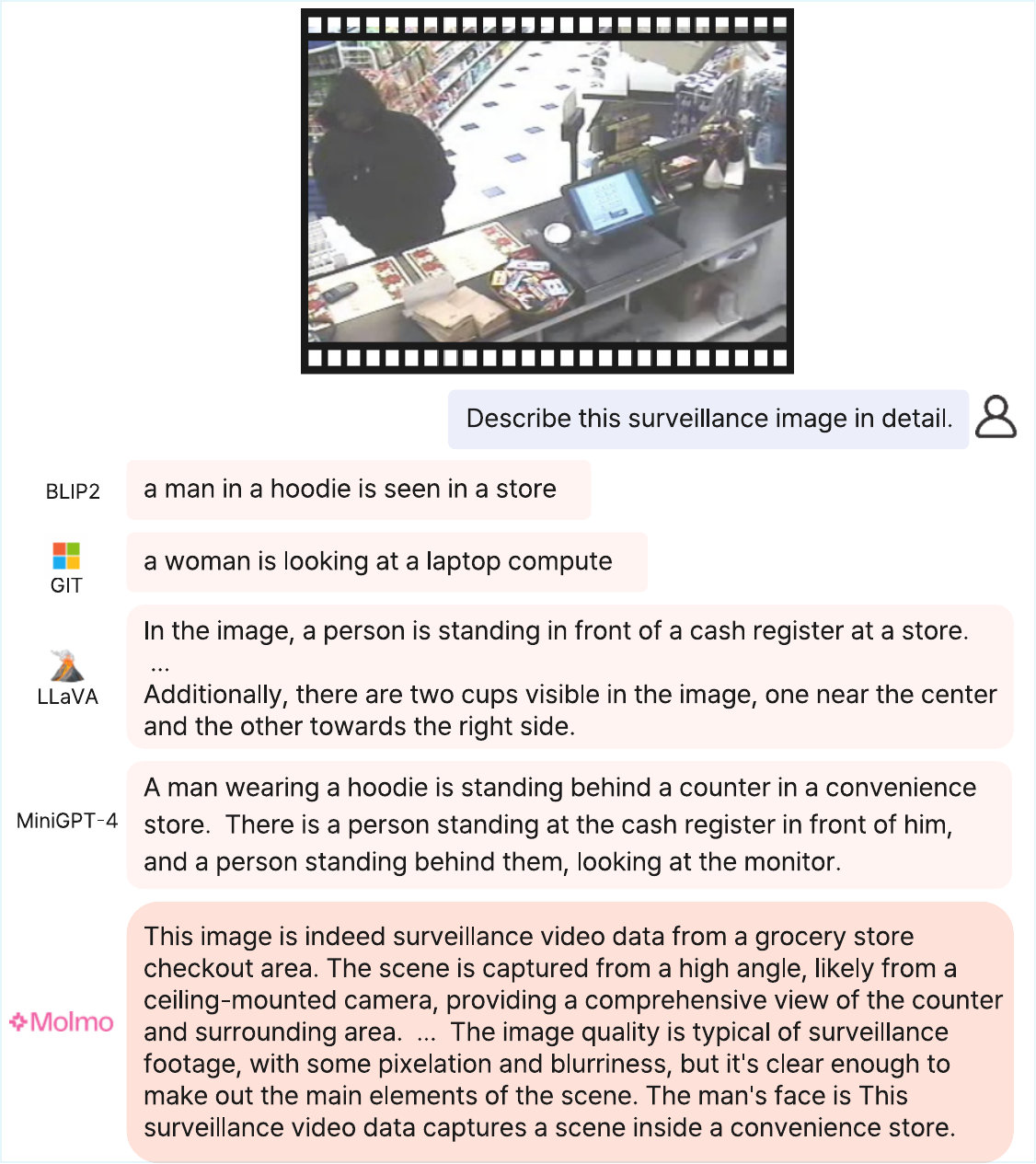}
  \caption{Comparison of captions generated by various VLMs. Molmo demonstrates the most descriptive and domain-appropriate outputs for surveillance imagery.}
    \label{fig:vlm}
\end{figure}

\begin{figure*}[ht!]
    \centering
    \includegraphics[width=1\linewidth]{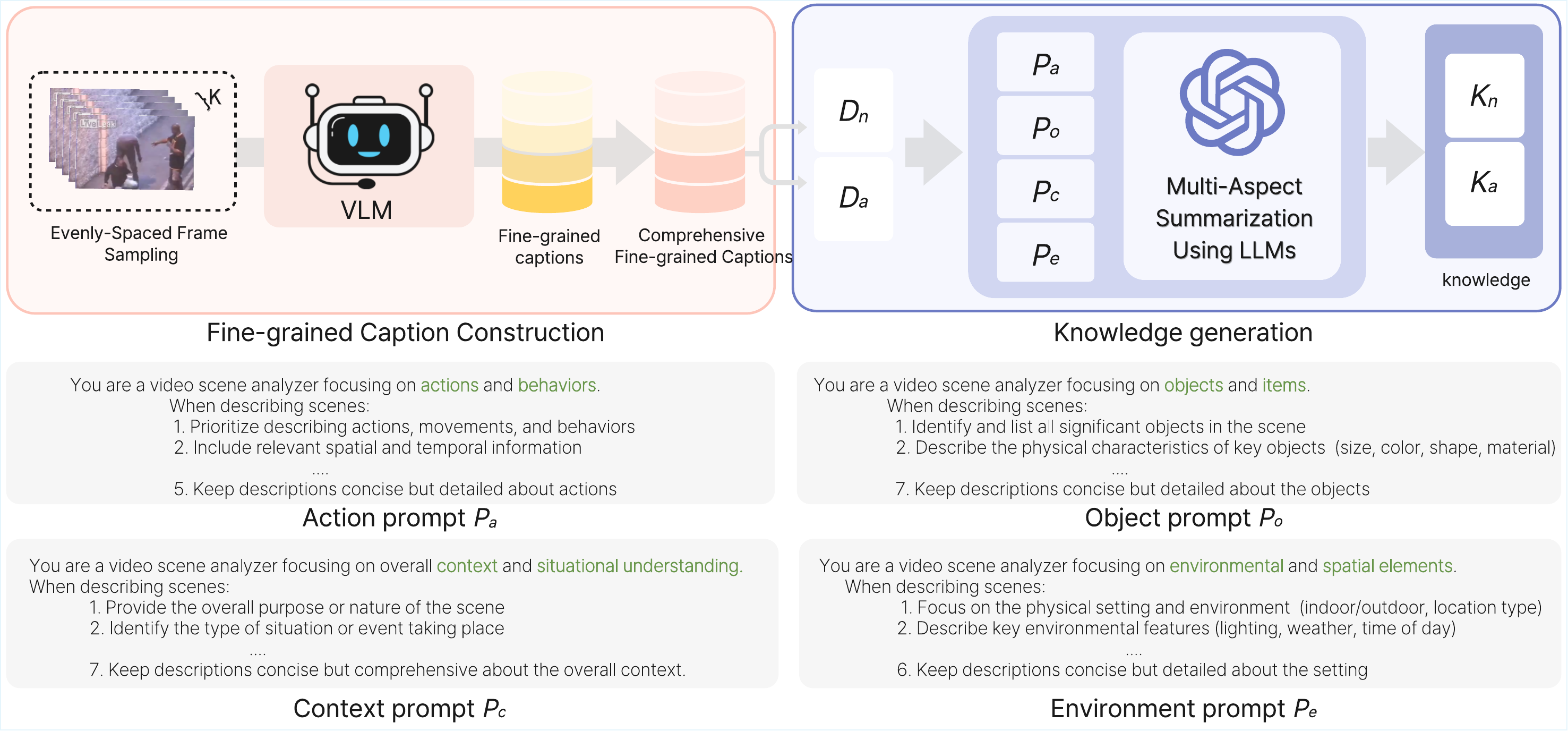}
    \caption{\textbf{Overview of knowledge generation.} Our pipeline first samples $K$ evenly spaced frames from a given video, and uses a frozen vision-language model (VLM) to generate fine-grained captions. These frame-level descriptions are aggregated across the video to form comprehensive textual summaries, grouped by video class: $D_n$ for normal videos and $D_a$ for abnormal ones. Using a large language model (LLM), we perform multi-aspect summarization with four prompts—$P_c$, $P_a$, $P_o$, and $P_e$—designed to extract context, action, object, and environmental information, respectively. The resulting structured knowledge is represented as $K_n$ and $K_a$, which encode multi-dimensional textual features of normal and abnormal scenarios. This process transforms raw visual content into interpretable and structured semantic representations for downstream anomaly detection.
    }
    \label{fig:gengeration_knowledge}
\end{figure*}

\subsection{Multimodal Anomaly Detection}
While most VAD methods rely solely on visual features, recent works have explored incorporating additional modalities—such as audio and text—to improve robustness and semantic understanding, particularly in low-resolution surveillance environments where visual signals are often ambiguous or incomplete. These multimodal approaches aim to compensate for the inherent limitations of vision-only models by introducing semantically richer or temporally complementary information streams.

Wu et al.~\cite{Wu} proposed a framework that fuses audio and visual streams, showing that auditory cues enhance recognition in complex scenarios. TPWNG~\cite{textprompt} introduced category-level textual prompts to guide visual representation learning, aligning semantic priors with video content for improved anomaly localization. TEVAD~\cite{TEVAD} further advanced this direction by leveraging SwinBERT~\cite{swinbert} to generate frame-level captions, which are combined with visual features to capture fine-grained semantic information. Their dual-captioning mechanism enabled better modeling of subtle anomalies by enriching temporal visual features with linguistic cues. Similarly, the UCA~\cite{Yuan2024uca} multimodal baseline augments visual inputs with caption-derived semantics for improved detection, reinforcing the potential of text-augmented models in surveillance settings.

These approaches demonstrate the benefits of using language to complement vision, enabling models to infer higher-level concepts such as intent, causality, and abnormal context. However, in all cases, text is treated as an auxiliary signal. In contrast, our work departs from this trend by investigating whether language alone—derived from frame-level captions and structured knowledge—can serve as the sole modality for anomaly detection, eliminating the need for direct visual input and offering a more interpretable, scalable, and modality-light alternative.

\subsection{Leveraging LLMs for VAD}

Recent studies have explored the use of Large Language Models (LLMs)\cite{llama,gpt4technicalreport,llmsurvey} in video anomaly detection to enhance semantic reasoning and interpretability. By leveraging their strong language understanding capabilities, LLMs enable the abstraction of complex visual events into structured textual formats, improving contextual understanding of anomalies and enabling higher-level inference. For example, LLMs can describe actions, object interactions, and event causality in human-readable forms, which are difficult to extract using visual features alone.

Prior works have leveraged LLMs for generating semantic summaries, guiding multiple instance learning, and facilitating causal reasoning about anomalous events~\cite{Zanella, Wu, Chen, Du}. These approaches demonstrate the potential of language-guided representations to serve as scalable and generalizable alternatives to traditional visual-based models—particularly in surveillance scenarios where visual signals are often ambiguous, occluded, or low in resolution. Moreover, incorporating language enhances the interpretability of detection results, which is critical in security-sensitive applications.

While these studies incorporate LLMs as auxiliary components alongside visual features, our work departs from this paradigm by fully eliminating the visual modality. We treat VAD as a purely textual reasoning task, using structured and instance-level descriptions as the sole input for detection. This text-only approach repositions LLMs not as mere enhancers of vision, but as central agents capable of understanding and detecting anomalies based solely on language-based representations.

\section{Methods}
\label{sec:methods}
Our proposed framework consists of three main components: (1) a Structured Knowledge Branch for constructing multi-aspect textual priors, (2) a Text Understanding Branch for encoding fine-grained captions, and (3) an Explainable Reasoning Branch for knowledge-grounded interpretability in anomaly detection.

\begin{figure*}[t]
    \centering
    \includegraphics[width=0.98\linewidth]{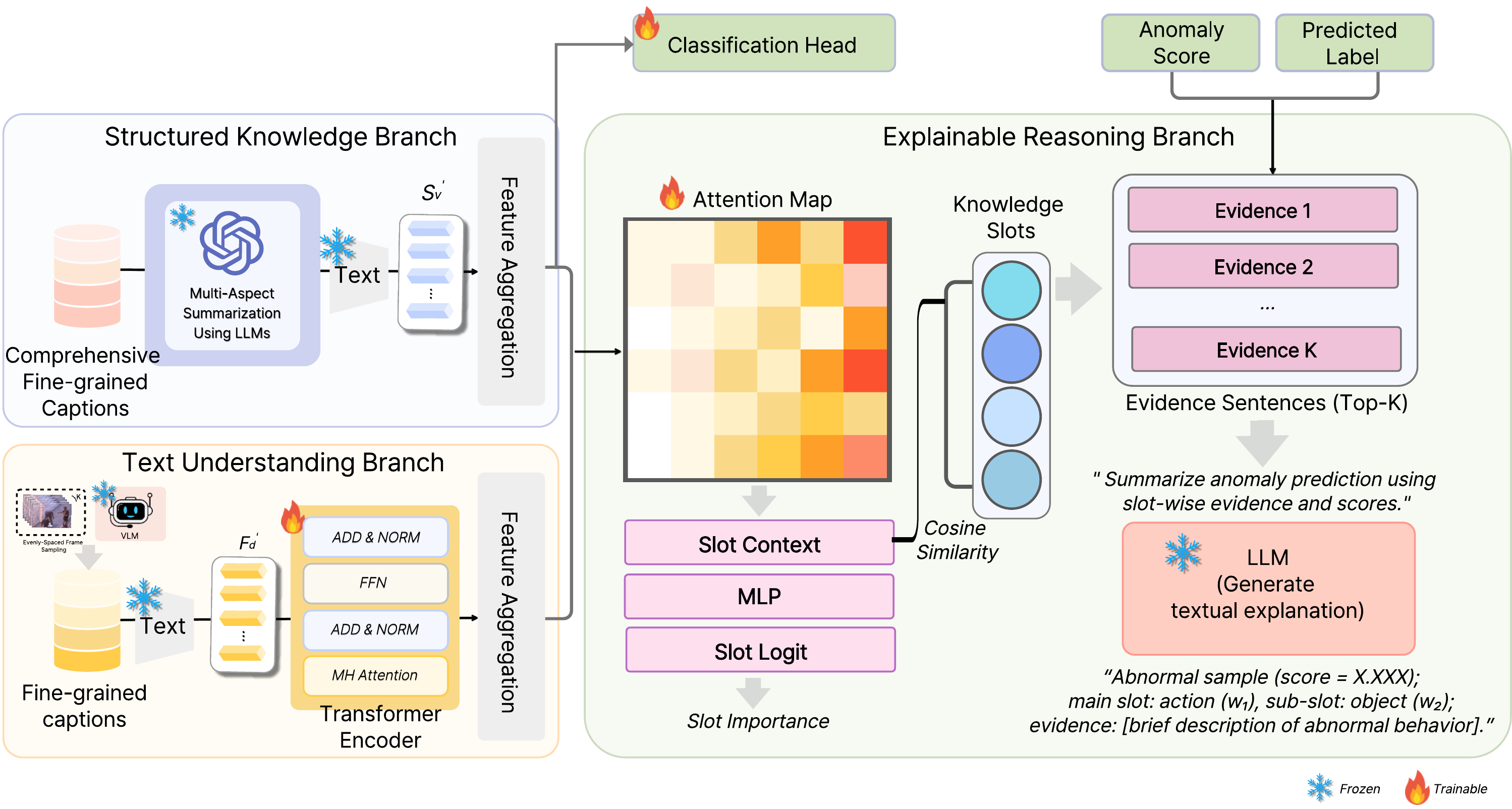}
    \caption{\textbf{Overview of the proposed TbVAD architecture.}
The framework consists of three main branches: (1) a Text Understanding Branch that encodes frame-level fine-grained captions using a trainable transformer encoder,
(2) a Structured Knowledge Branch that constructs multi-aspect textual priors (context, action, object, environment) from video captions via LLM-based summarization, and
(3) an Explainable Reasoning Branch that computes slot-wise importance through attention-based alignment and generates human-interpretable textual explanations.
The representations from the first two branches are fused through a classification head to predict the anomaly score and label,
while the reasoning branch provides slot-grounded evidence supporting each prediction.}
    \label{fig:architecture}
\end{figure*}

\subsection{Structured Knowledge Branch}

\subsubsection{ Overview and Motivation}
To detect anomalies using textual semantics, our framework constructs structured knowledge by transforming visual inputs into rich language-based representations. This process begins with extracting fine-grained captions from sampled frames and proceeds through multi-perspective summarization using a Large Language Model (LLM)\cite{gpt4technicalreport}.

\subsubsection{Benchmarking Vision-Language Models for Captioning}
Given the low resolution and cluttered nature of surveillance footage, selecting an effective Vision-Language Model (VLM) for caption generation is crucial.
We empirically evaluated five representative VLMs—BLIP2~\cite{blip2}, GIT~\cite{git}, LLaVA~\cite{llava}, MiniGPT-4~\cite{minigpt4}, and Molmo~\cite{Molmo}—across diverse CCTV samples.
As shown in Fig.~\ref{fig:vlm}, Molmo consistently produced the most detailed and context-aware captions, outperforming other models that often generated short or generic phrases (e.g., “a man in a hoodie is seen in a store”).

To quantify this observation, Table~\ref{tab:vlm_table} reports the average caption length and TF-IDF score for each model.
Molmo yielded the longest and most semantically rich captions, effectively capturing object details, scene context, and action cues—features that are critical for anomaly reasoning.

Although Molmo demonstrated superior captioning capability,
we did not apply it to UCF-Crime due to its per-frame evaluation protocol,
which would require generating captions for every individual frame,
making large-scale inference computationally infeasible.
Instead, we utilized Molmo to construct structured knowledge representations used in the reasoning stage,
ensuring that its semantic understanding still contributes to the overall framework.
For XD-Violence, which employs video-level labels across untrimmed clips,
Molmo was adopted as the primary caption generator for fine-grained textual descriptions.

\subsubsection{Multi-Aspect Summarization via LLMs}

As illustrated in Fig.~\ref{fig:gengeration_knowledge}, we construct structured knowledge from fine-grained captions (i.e., aggregated across multiple frames to form comprehensive descriptions) through multi-aspect summarization using a Large Language Model (LLM)~\cite{gptturbo}. For each input video, \(K\) evenly spaced frames are sampled, and fine-grained captions are generated using a frozen Vision-Language Model (VLM)~\cite{Molmo}. The resulting captions are grouped by video labels into \(D_n\) for normal and \(D_a\) for abnormal samples.

\begin{table}[h]
\centering
\begin{tabular}{l c|c c}
\toprule
\textbf{Model} & \makecell[c]{\textbf{\#Trainable} \\ \textbf{Params}} & \textbf{Avg. Len} & \textbf{TF-IDF} \\
\midrule
GIT  &1B &9.73 & 1.9621\\
BLIP2 &2.7B & 10.52 &1.9829 \\
MiniGPT-4  &13B&49.09 &3.2602 \\
LLaVA  &7B & 91.28 & 1.9526\\
\rowcolor[HTML]{EFEFEF} Molmo &7B& \textbf{2602.86}& \textbf{4.2929}       \\
\bottomrule
\end{tabular}
\caption{Comparison of average caption length and information content (TF-IDF) for different models on the UCF dataset. All captions were generated on the same video set to ensure fair comparison.}
\label{tab:vlm_table}
\end{table}

To extract structured semantics, we design four tailored prompts targeting distinct aspects of each scene: context (\(P_c\)), action (\(P_a\)), object (\(P_o\)), and environment (\(P_e\))—as illustrated in Fig.~\ref{fig:gengeration_knowledge}. Each prompt is applied to both the normal and abnormal caption groups. Let \(V = \{n, a\}\) represent the set of normal and abnormal labels. For each aspect \(K \in \{C, A, O, E\}\), we obtain summaries \(K_v\), where \(v \in V\).
These elements constitute the structured knowledge for each class: \(K_v = \{C_v, A_v, O_v, E_v\}\) for \(v \in V\). 
The complete representation is thus defined as \(K = \{K_n, K_a\}\). This structured representation captures rich semantic priors for both normal and abnormal events, enabling the model to identify anomalies based on text even when visual signals are ambiguous or noisy.

\subsubsection{ Knowledge Encoding}
This process takes the structured representation \(K = \{C, A, O, E\}\), composed of four textual components of video. These components are concatenated and encoded as a single sequence using a frozen language model, producing token embeddings \(S_V^{'} = \{s_1, s_2, \dots, s_L\}\). The final knowledge representation is obtained by projecting the mean of these embeddings into a shared latent space: 
 
\begin{equation}
P_V = W_V \cdot \left( \frac{1}{L} \sum_{i=1}^{L} s_i \right) + b_V
\end{equation}

\subsection{ Text Understanding Branch}

\subsubsection{Fine-Granined Captions Encoding}

Given a video, we sample \(K\) evenly spaced frames and generate corresponding captions \(F_d^{'} = \{c_1, c_2, \dots, c_K\}\) using a frozen vision-language model. These captions are tokenized and embedded into vectors \(X_d = \{x_1, x_2, \dots, x_K\}\), which serve as the initial input to the Transformer encoder, i.e., \(Z^{(0)} = X_d\).

The sequence is then processed through a stack of \(L\) Transformer encoder layers:

\begin{equation}
Z^{(l)} = \text{Encoder}^{(l)}(Z^{(l-1)}), \quad l = 1, \dots, L
\end{equation}

The final output \(H_d = Z^{(L)} = \{h_1, h_2, \dots, h_K\}\) is aggregated via average pooling and projected into the latent space:

\begin{equation}
P_d = W_d \cdot \left( \frac{1}{K} \sum_{i=1}^{K} h_i \right) + b_d
\end{equation}

\subsubsection{ Feature Aggregation and Classification}

The two projected vectors \(P_d\) and \(P_V\) are concatenated through a \textit{Feature Fusion Module} and passed through a \textit{Classification Head} to produce the anomaly probability:
\begin{equation}
    y = \sigma(W [P_d; P_V] + b)
\end{equation}

Here, \(y\) denotes the predicted probability of an anomaly. This architecture effectively combines detailed frame-level semantics with high-level contextual knowledge, enabling robust and interpretable video anomaly detection.
\begin{table*}[h]
    \centering
    \begin{tabular}{clccc}
        \toprule
        \multirow{2}{*}{\textbf{}} &\multirow{2}{*}{\textbf{Methods}}  &\multirow{2}{*}{\textbf{Modality}}  & \textbf{UCF} & \textbf{XD} \\ 
        &  &  & AUC (\%) & AP (\%) \\
        \midrule
        \multirow{14}{*}{{\rotatebox[origin=c]{90}{\textbf{Weakly-supervised}} }} 
        & Sultani et al. \cite{sultani} & V & 77.92 & 73.20 \\
        & GCN \cite{zhong} & V & 82.12 & - \\
        & CLAWS \cite{claws} & V & 82.30 & - \\
        & MIST \cite{MIST} & V & 82.30 & - \\
        & RTFM \cite{tian} &  V & 84.30 & 77.81 \\ 
        & CRFD \cite{CRFD} &  V & 84.89 & 75.90 \\
        & GCL \cite{GCL}  & V & 79.84 & - \\
        & MSL \cite{MSL}  & V & 85.62 & 78.58 \\
        & MGFN \cite{MGFN} & V & 86.67 & 80.11 \\
        & Zhang et al. \cite{Zhang} & V  & 86.22 & 78.74 \\
        & Wu et al.\cite{open} & V & 86.40 & 66.53 \\
        & UR-DMU \cite{DMU} &  V & 86.97 & 81.66 \\
        & UMIL \cite{MIL} & V & 86.75 & - \\
        & CLIP-TSA \cite{TSA}  & V & 87.58 & 82.17 \\
        & TPWNG \cite{textprompt} & V+C & 87.79 & 83.68 \\
            \rowcolor[HTML]{EFEFEF}
        & \textbf{Our TbVAD w GeneratedCap} & T  & \textbf{85.42} & \textbf{97.34} \\

        \bottomrule
    \end{tabular}
    \caption{Performance comparison on UCF and XD datasets. \textbf{Best results} are shown in bold. 
The modality column specifies the input type used in each approach. "V", "T", and "C" represent visual, text, and category label data, respectively.}
    \label{tab:performance}
\end{table*}
\subsection{Explainable Reasoning Branch}
\label{sec:explainable_branch}

\subsubsection{Slot-wise Importance Estimation}

Given the description feature matrix $H_d \in \mathbb{R}^{T\times d}$ 
and the knowledge slot prototypes $K_v \in \mathbb{R}^{S\times d}$ 
($S{=}4$, corresponding to \textit{context}, \textit{action}, \textit{object}, and \textit{environment}), 
we first compute a cross-attention map that aligns each slot with the most relevant tokens in the description:

\begin{equation}
A = \frac{K_v H_d^\top}{\sqrt{d}}, \qquad
C = A H_d.
\end{equation}

Here, $A \in \mathbb{R}^{S\times T}$ represents the attention alignment between slots and tokens, 
and $C \in \mathbb{R}^{S\times d}$ denotes the resulting slot-specific context vectors, 
obtained as weighted combinations of the token embeddings.

Each slot’s importance is then estimated by fusing its context vector and prototype embedding 
through a lightweight projection network:

\begin{equation}
z_s = f([C_s; K_{v,s}]), \qquad
w_s = \frac{\exp(z_s)}{\sum_{j=1}^{S} \exp(z_j)}.
\end{equation}

Here, $f(\cdot)$ denotes a two-layer feed-forward transformation, and $w_s$ 
represents the normalized importance of slot $s$.
Higher $w_s$ values indicate stronger contributions to the anomaly decision.

\subsubsection{Knowledge-grounded Evidence Retrieval}

For each slot, we identify the most semantically aligned knowledge sentence 
by measuring cosine similarity between the mean description embedding 
and the slot-specific knowledge representations:

\begin{equation}
e^{\star}_{v,s} 
= \arg\max_{e \in E_{v,s}} 
\frac{\bar h^\top e}{\|\bar h\|_2 \|e\|_2},
\end{equation}

where $\bar h$ denotes the mean-pooled description representation, 
and $E_{v,s}$ is the set of candidate knowledge sentences for slot $s$ under class $v$.
Among all slots, the top-$k$ (typically $k{=}2$) are retained as supporting evidence, 
providing interpretable textual grounding for the model’s decision.

\subsubsection{Textual Explanation Generation}

Finally, the retrieved evidences, slot importance scores, and predicted label 
are combined into a structured record $R$.  
An explanation generator module then produces a concise natural-language justification 
based on $R$, summarizing the primary factors that contributed to the anomaly decision.  

This generated explanation reflects the internal reasoning process of the model, 
linking quantitative predictions with interpretable textual evidence 
to enhance the transparency and reliability of the detection outcome.

\section{Experiments}
\label{sec:experiments}

Our TbVAD framework employs Longformer to encode structured knowledge (up to 4096 tokens) and BERT to encode fine-grained textual descriptions (up to 512 tokens), providing efficient representations for both global and local semantics.  

Structured knowledge is constructed using GPT-3.5 Turbo, 
which organizes event semantics into four predefined slots: 
\textit{context}, \textit{action}, \textit{object}, and \textit{environment}.  
Fine-grained captions are generated by the vision-language model Molmo, 
which produces detailed frame-level descriptions from surveillance videos.  

Finally, the explanation module utilizes the instruction-tuned large language model Qwen2.5-7B-Instruct to generate concise, human-understandable rationales 
that align with the model’s textual reasoning process.

\subsection{Dataset}
\textbf{UCF-Crime.} The UCF-Crime dataset \cite{ucfc} is a large-scale benchmark for video anomaly detection, containing 1,900 untrimmed surveillance videos across 13 real-world anomaly categories such as robbery, fighting, and arson.\\ 
\textbf{UCA.} The UCA dataset \cite{Yuan2024uca} is an extension of UCF-Crime that additionally offers human-written sentence-level descriptions for each video. In our experiments, we use both the video-level anomaly labels from UCF-Crime and the descriptive annotations from UCA\cite{Yuan2024uca} to support text-based anomaly detection.\\
\textbf{XD-Violence.} The XD-Violence dataset \cite{xd} contains 4,754 untrimmed surveillance videos with a total duration of 217 hours. It covers a wide range of violent and non-violent events such as abuse, explosions, and riots, captured across multiple real-world scenes.\\

\subsection{Performance Metric}
We evaluate our models on the UCA\cite{Yuan2024uca} and XD\cite{xd} datasets. Performance is measured using Area Under the ROC Curve (AUC) for UCA\cite{Yuan2024uca}, which reflects the model's ability to distinguish between normal and abnormal events across various thresholds. For XD \cite{xd}, we use Average Precision (AP), which summarizes the precision-recall trade-off and is well-suited for imbalanced datasets. To assess generalization capability, we additionally conduct cross-domain evaluations by training on one dataset and testing on the other.

\subsection{Quantitative Results}

Table~\ref{tab:performance} compares our TbVAD framework with state-of-the-art (SOTA) methods on the UCF-Crime and XD-Violence datasets.  
Unlike conventional approaches that depend on visual representations, TbVAD performs anomaly detection entirely within the textual domain, relying on generated captions and structured knowledge representations.  

The proposed TbVAD w GeneratedCap variant utilizes automatically generated captions—BLIP2-based descriptions for UCF-Crime and Molmo-generated captions for XD-Violence.  
This design enables TbVAD to assess anomalies from textual semantics alone, without using any visual input.  
As shown in Table~\ref{tab:performance}, TbVAD achieves competitive or superior performance to visual-feature-based baselines across both datasets, demonstrating the effectiveness of text-only reasoning for anomaly detection.  

The results further highlight that caption quality and dataset characteristics play crucial roles: XD-Violence videos are typically clearer and captured at closer viewpoints, allowing vision-language models to generate highly descriptive and accurate captions. In contrast, UCF-Crime videos often contain low-resolution or distant scenes, which limit the precision of automatically generated captions. These findings indicate that the success of language-based VAD depends strongly on the expressiveness and contextual richness of the underlying textual inputs.

\begin{table}[h]
\centering
\begin{tabular}{cccc|cc}
\toprule
\multicolumn{4}{c}{\textbf{structure component}} & \textbf{UCF (AUC)} & \textbf{XD (AP)} \\
\textbf{ac} & \textbf{ob} & \textbf{co} & \textbf{en} & AUC(\%) & AP(\%) \\
\midrule
\checkmark & & \checkmark & &76.98&97.62\\
\checkmark & & & & 85.10 & 96.99 \\
\checkmark & \checkmark &         &         & 84.36 & 97.45 \\
\checkmark & \checkmark & \checkmark &         & 85.33 & 96.89 \\
\checkmark & &  & \checkmark & 82.27 & 96.57 \\
\checkmark & \checkmark & \checkmark & \checkmark & 83.42 & 97.03 \\
\rowcolor[HTML]{EFEFEF} & \checkmark & & \checkmark & \textbf{85.42} & \textbf{97.65} \\
\bottomrule
\end{tabular}
\caption{Performance of selected knowledge combinations on UCF and XD datasets. 
A \checkmark~denotes the inclusion of a knowledge component. 
Here, \textit{ac}, \textit{ob}, \textit{co}, and \textit{en} represent \textit{action}, \textit{object}, \textit{context}, and \textit{environment}, respectively.}
\label{tab:structure}
\end{table}

\begin{table*}[h]
\centering
\renewcommand{\arraystretch}{1.3}
\begin{tabular}{c|cc|cc||cc|cc}
\toprule
& \multicolumn{4}{c||}{Prior Work\cite{open}} 
& \multicolumn{4}{c}{\textbf{Ours}} \\
\cline{2-9}
Test$\Rightarrow$
& \multicolumn{2}{c|}{UCF} & \multicolumn{2}{c||}{XD} 
& \multicolumn{2}{c|}{UCF} & \multicolumn{2}{c}{XD} \\
Train$\Downarrow$
& AUC(\%) & ACC(\%) & AP(\%) & ACC(\%) 
& AUC(\%) & ACC(\%) & AP(\%) & ACC(\%) \\
\hline
UCF & \cellcolor[HTML]{EFEFEF}\textbf{86.05} & \cellcolor[HTML]{EFEFEF}\textbf{45.00} & 63.74 & 47.90 
                & \cellcolor[HTML]{EFEFEF}\textbf{85.56} & \cellcolor[HTML]{EFEFEF}\textbf{77.59} & 74.93 & 67.00 \\
XD     & 82.42 & 40.71 & \cellcolor[HTML]{EFEFEF}\textbf{82.86} & \cellcolor[HTML]{EFEFEF}\textbf{88.96}
                & 63.59 & 66.90 & \cellcolor[HTML]{EFEFEF}\textbf{98.00} & \cellcolor[HTML]{EFEFEF}\textbf{90.25} \\
\bottomrule
\end{tabular}
\caption{ Cross-Dataset Results on UCF and XD.}
\label{tab:cross_domain}
\end{table*}

\begin{figure}[t]
    \centering

    \begin{subfigure}[b]{1\linewidth}
        \centering
        \includegraphics[width=\textwidth]{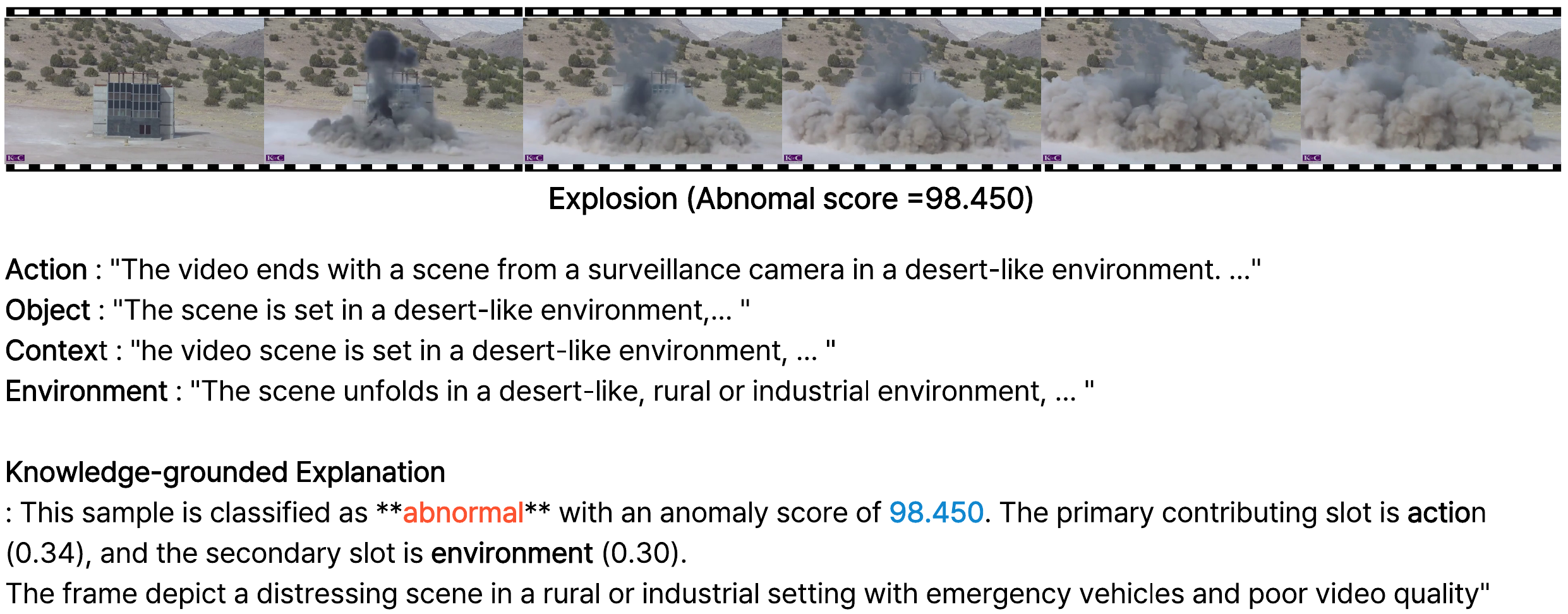}
        \caption{TbVAD-generated explanation for abnormal event: \textit{Explosion}. }
        \label{fig:a}
    \end{subfigure}

    \vspace{2mm} 

    \begin{subfigure}[b]{1\linewidth}
        \centering
        \includegraphics[width=\textwidth]{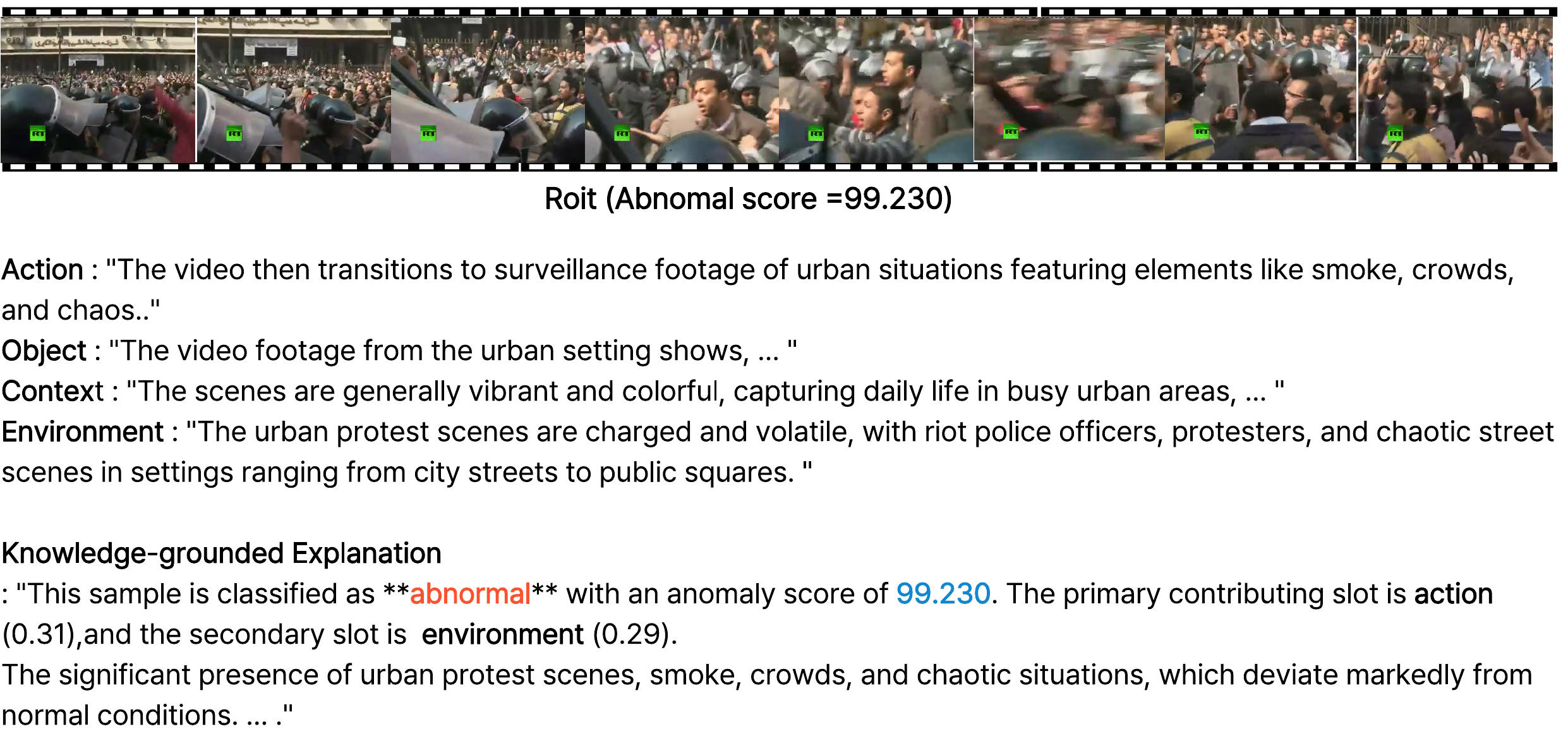}
        \caption{TbVAD-generated explanation for abnormal event: \textit{Riot}.}
        \label{fig:b}
    \end{subfigure}

    \vspace{2mm}

    \begin{subfigure}[b]{1\linewidth}
        \centering
        \includegraphics[width=\textwidth]{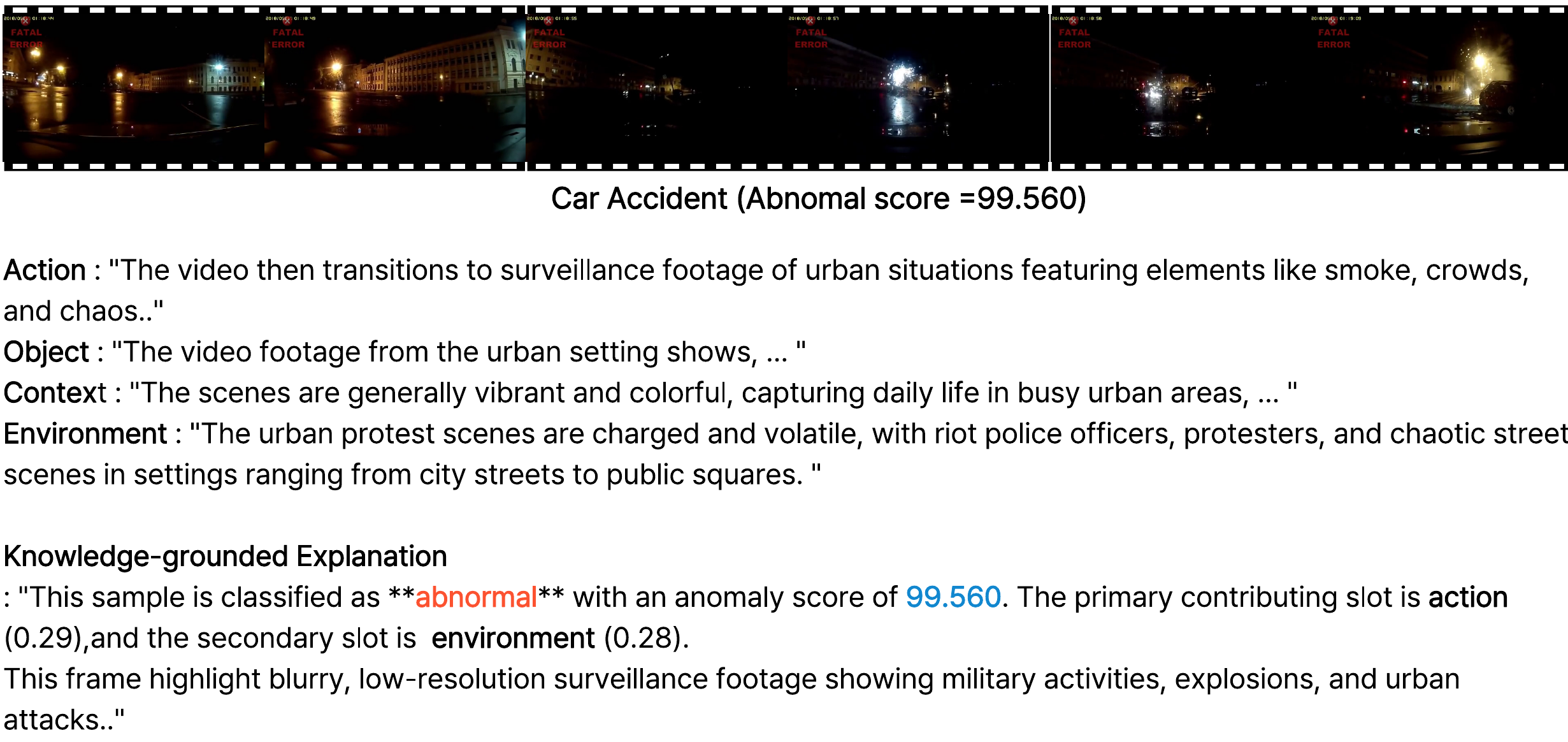}
        \caption{TbVAD-generated explanation for abnormal event: \textit{Car Accident}.}
        \label{fig:c}
    \end{subfigure}

   \caption{
Visualization of TbVAD’s knowledge-grounded explanations for three representative abnormal events: 
(a) \textit{Explosion}, (b) \textit{Riot}, and (c) \textit{Car Accident}.
}

    \label{fig:qualitative}
\end{figure}

\begin{figure}[t]
    \centering
    \begin{subfigure}[b]{0.31\linewidth}
        \includegraphics[width=\linewidth]{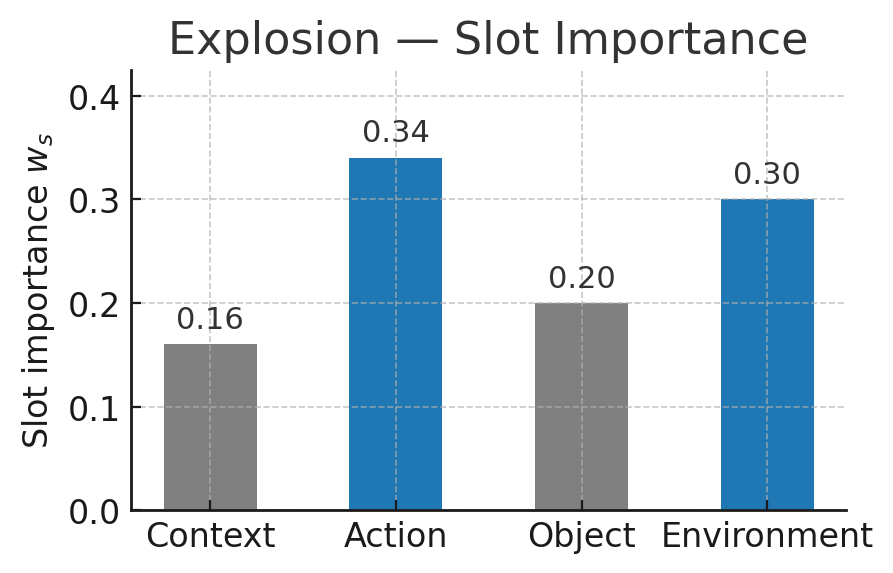}
        \caption{Explosion}
    \end{subfigure}
    \begin{subfigure}[b]{0.31\linewidth}
        \includegraphics[width=\linewidth]{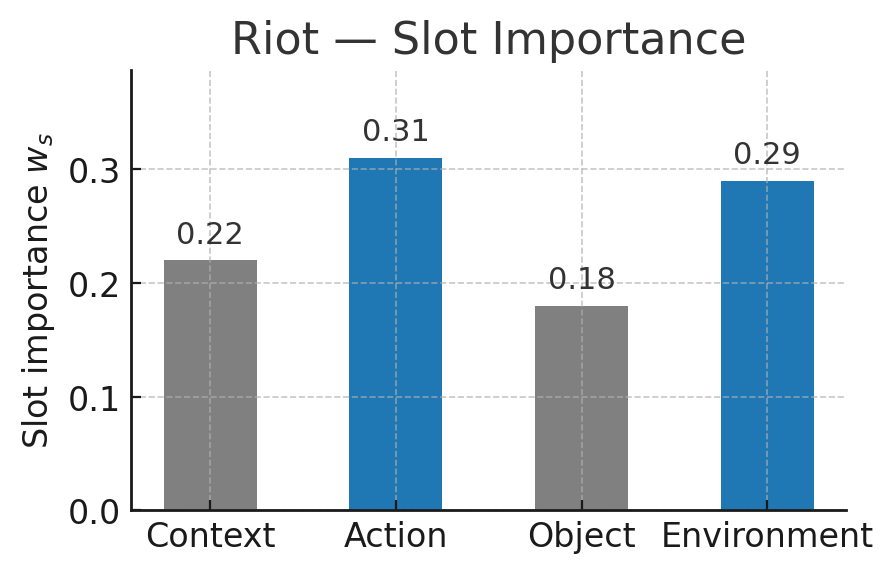}
        \caption{Riot}
    \end{subfigure}
    \begin{subfigure}[b]{0.31\linewidth}
        \includegraphics[width=\linewidth]{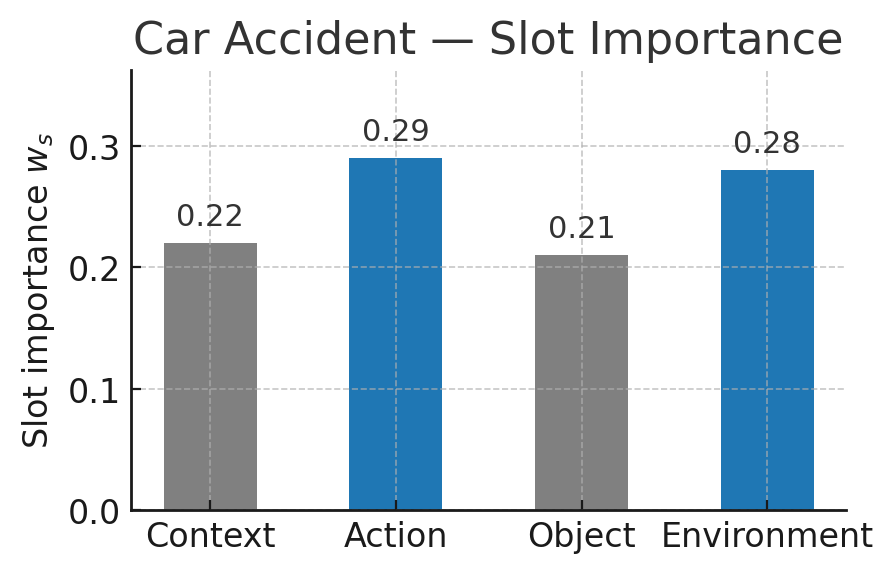}
        \caption{Car Accident}
    \end{subfigure}

    \vskip 2pt
    \begin{subfigure}[b]{0.31\linewidth}
        \includegraphics[width=\linewidth]{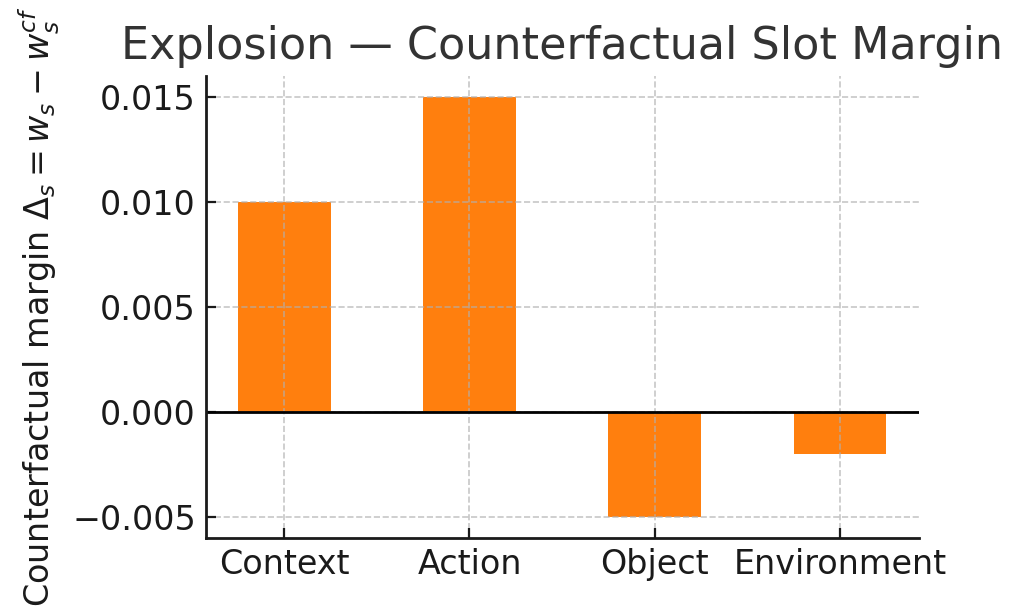}
        \caption{Explosion (Counterfactual)}
    \end{subfigure}
    \begin{subfigure}[b]{0.31\linewidth}
        \includegraphics[width=\linewidth]{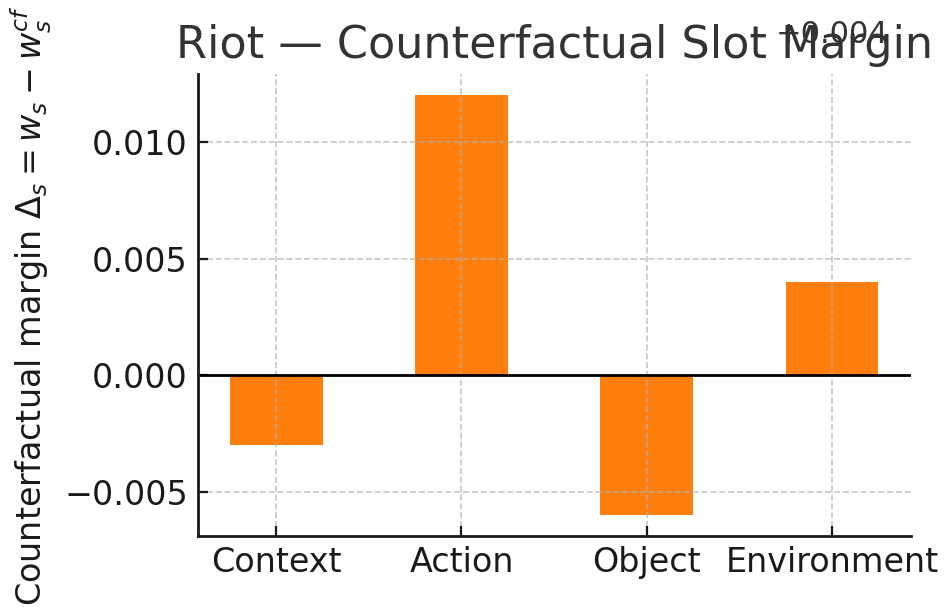}
        \caption{Riot (Counterfactual)}
    \end{subfigure}
    \begin{subfigure}[b]{0.31\linewidth}
        \includegraphics[width=\linewidth]{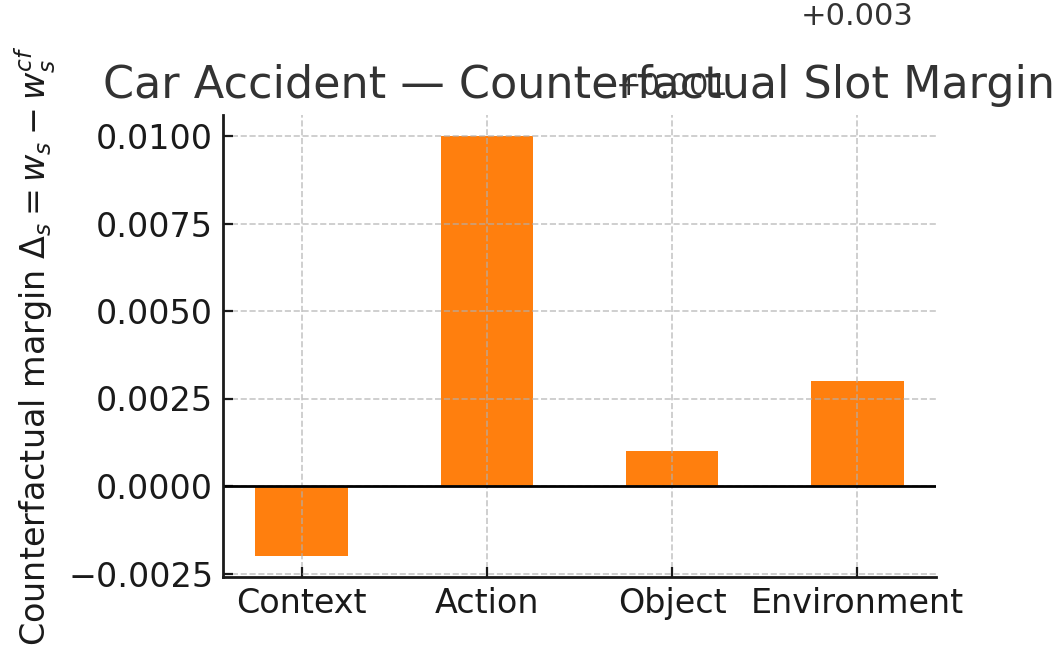}
        \caption{Car Accident (Counterfactual)}
    \end{subfigure}

    \caption{\textbf{Visualization of slot-wise importance and counterfactual reasoning in TbVAD.}
    (a–c) show slot importance ($w_s$) for Explosion, Riot, and Car Accident samples, respectively.
    (d–f) visualize the corresponding counterfactual slot margins ($\Delta_s = w_s - w^{cf}_s$),
    illustrating how each slot’s contribution changes when the prediction is inverted.
    Blue bars denote the top-2 influential slots; gray bars represent minor ones.}
    \label{fig:slot}
\end{figure}

\subsection{Structured Knowledge Composition Analysis}

We decompose structured knowledge into four semantically distinct dimensions: \textit{action}, \textit{object}, \textit{context}, and \textit{environment}. These components reflect the key elements of surveillance scenes and serve as the basis for our structured representation.
To explore how each dimension contributes to anomaly detection, we perform an ablation study on different combinations. Table~\ref{tab:structure} summarizes the results. Interestingly, certain combinations—such as \textit{object} + \textit{environment}—achieve higher accuracy than using all four components together. For example, on the UCF dataset, this pair reaches an AUC of 85.42\%, surpassing the full combination 83.42\%. In contrast, the lowest performance 76.98\% is observed when only \textit{action} and \textit{context} are used.

These findings demonstrate the flexibility and modularity of our structured knowledge design. Depending on the dataset characteristics, different semantic dimensions contribute more significantly to anomaly detection. While Table~\ref{tab:performance} confirms that using the full combination still yields strong overall performance, the ablation results suggest that selective integration of components may offer further room for optimization. This compositional structure thus supports adaptable and interpretable anomaly reasoning under varying conditions.

\subsection{Cross-Dataset Evaluation}

To evaluate the generalization ability of TbVAD, we conduct cross-domain experiments by training on one dataset and testing on another. Table~\ref{tab:cross_domain} presents the results of these experiments, where the model is trained on UCF-Crime~\cite{ucfc} and evaluated on XD-Violence~\cite{xd}, and vice versa. 

The results indicate that incorporating structured knowledge improves cross-dataset performance, suggesting that learning generalizable semantic priors helps the model adapt beyond the distribution of the training data. This highlights the potential of knowledge-guided approaches in enhancing domain robustness in video anomaly detection.

\subsection{Explainable Reasoning Analysis}

Figure~\ref{fig:qualitative} and Figure~\ref{fig:slot} present qualitative examples illustrating how TbVAD performs explainable reasoning through structured textual understanding.  
In Figure~\ref{fig:qualitative}, we visualize three representative abnormal samples—\textit{Explosion}, \textit{Riot}, and \textit{Car Accident}—along with their corresponding slot-wise analyses and generated explanations.  
For each case, TbVAD identifies the most influential semantic slots (e.g., \textit{action}, \textit{environment}) that primarily contribute to the abnormality decision and retrieves the most relevant textual evidences from structured knowledge to justify the prediction.  
The resulting explanations provide human-readable insights into why the event was classified as abnormal, confirming that the model’s reasoning process aligns with intuitive human understanding.

Figure~\ref{fig:slot} further visualizes slot importance and counterfactual reasoning results.  
Subfigures (a–c) show the relative importance of each slot, where the top-2 dominant slots (colored in blue) correspond to the key semantic factors influencing the decision.  
Subfigures (d–f) display the corresponding counterfactual slot margins, indicating how each slot’s contribution changes when the prediction is inverted.  
A positive margin suggests that reducing the influence of a specific slot (e.g., \textit{action}) could lead the model to interpret the event as normal.  
These visualizations demonstrate that TbVAD not only detects anomalies through textual semantics but also provides interpretable, knowledge-grounded reasoning for its predictions.
\section{Conclusion}
\label{sec:conclusion}

We introduced TbVAD, a novel text-based framework for Video Anomaly Detection (VAD) that operates entirely within the language domain by leveraging fine-grained captions and structured knowledge.
By transforming surveillance videos into rich textual representations, TbVAD enables robust and interpretable anomaly detection under weak supervision.
Our framework demonstrates strong performance on UCF-Crime and XD-Violence, validating the effectiveness of textual reasoning for understanding complex events in real-world scenarios.

Beyond detection accuracy, TbVAD offers practical advantages in scalability and efficiency, removing dependence on high-quality visual inputs.
Future work will focus on extending TbVAD toward real-time analysis and enhancing the generality of its explanations across diverse event types.
By identifying not only anomalies but also their semantic precursors, TbVAD holds promise for proactive safety monitoring and intelligent surveillance applications.
{
    \small
    \bibliographystyle{ieeenat_fullname}
    \bibliography{main}
}


\end{document}